\documentclass[conference]{IEEEtran}

\makeatletter

\def\ps@IEEEtitlepagestyle{%
  \def\@oddfoot{\mycopyrightnotice}%
  \def\@evenfoot{}%
}
\def\mycopyrightnotice{%
  {\footnotesize  979-8-3315-7191-7/26\$31.00~\copyright~2026 IEEE\hfill}% <--- Change here
  \gdef\mycopyrightnotice{}
}

\usepackage{blindtext}
\usepackage{eso-pic}
\IEEEoverridecommandlockouts
\usepackage{cite}
\usepackage{amsmath,amssymb,amsfonts}
\usepackage{algorithmic}
\usepackage{graphicx}
\usepackage{textcomp}
\usepackage{xcolor}
\usepackage{multicol}
\usepackage{hyperref}
\usepackage{cleveref}
\usepackage{multirow}
\usepackage{makecell}
\usepackage{arydshln}
\def\BibTeX{{\rm B\kern-.05em{\sc i\kern-.025em b}\kern-.08em
    T\kern-.1667em\lower.7ex\hbox{E}\kern-.125emX}}
    
\usepackage{eso-pic}
\newcommand\AtPageUpperMyright[1]{\AtPageUpperLeft{%
 \put(\LenToUnit{0.17\paperwidth},\LenToUnit{-2cm}){%
     \parbox{0.9\textwidth}{\raggedleft\fontsize{8}{11}\selectfont #1}}%
 }}%
\newcommand{\conf}[1]{%
\AddToShipoutPictureBG*{%
\AtPageUpperMyright{#1}
}
}

\begin{document}
\title{\vspace*{1cm} Diverse Word Choices, Same Reference: Annotating Lexically-Rich Cross-Document Coreference 
}

\author{\IEEEauthorblockN{Anastasia Zhukova}
\IEEEauthorblockA{
% \textit{dept. name of organization (of Aff.)} \\
\textit{University of Göttingen}\\
Göttingen, Germany \\
0000-0001-9084-2890}
\and
\IEEEauthorblockN{Felix Hamborg}
\IEEEauthorblockA{
% \textit{dept. name of organization (of Aff.)} \\
\textit{Humboldt University}\\
Berlin, Germany \\
0000-0003-2444-8056}
\and
\IEEEauthorblockN{Karsten Donnay}
\IEEEauthorblockA{
% \textit{dept. name of organization (of Aff.)} \\
\textit{University of Zurich}\\
Zurich, Switzerland \\
0000-0002-9080-6539}
\and
\IEEEauthorblockN{Norman Meuschke}
\IEEEauthorblockA{
% \textit{dept. name of organization (of Aff.)} \\
\textit{University of Göttingen}\\
Göttingen, Germany \\
0000-0003-4648-8198}
\and
\IEEEauthorblockN{Bela Gipp}
\IEEEauthorblockA{
% \textit{dept. name of organization (of Aff.)} \\
\textit{University of Göttingen}\\
Göttingen, Germany \\
0000-0001-6522-3019}
% \and
% \IEEEauthorblockA{
% % \textit{dept. name of organization (of Aff.)} \\
}

\maketitle
\conf{\textit{  Proc. of International Conference on Artificial Intelligence, Computer, Data Sciences and Applications (ACDSA 2026) \\ 
5-7 February 2026, Boracay-Philippines}}

\begin{abstract}
Cross-document coreference resolution (CDCR) identifies and links mentions of the same entities and events across related documents, enabling content analysis that aggregates information at the level of discourse participants. However, existing datasets primarily focus on event resolution and employ a narrow definition of coreference, which limits their effectiveness in analyzing diverse and polarized news coverage where wording varies widely. This paper proposes a revised CDCR annotation scheme of the NewsWCL50 dataset, treating coreference chains as discourse elements (DEs) and conceptual units of analysis. The approach accommodates both identity and near-identity relations, e.g., by linking ``the caravan'' - ``asylum seekers'' - ``those contemplating illegal entry'', allowing models to capture lexical diversity and framing variation in media discourse, while maintaining the fine-grained annotation of DEs. We reannotate the NewsWCL50 and a subset of ECB+ using a unified codebook and evaluate the new datasets through lexical diversity metrics and a same-head-lemma baseline. The results show that the reannotated datasets align closely, falling between the original ECB+ and NewsWCL50, thereby supporting balanced and discourse-aware CDCR research in the news domain.
\end{abstract}

% \mycopyrightnotice{979-8-3315-7191-7/26/\$31.00 ©2026 IEEE}


\section{Introduction}
Cross-document coreference resolution (CDCR) is a Natural Language Processing (NLP) task that aims to identify and link mentions of the same entities and events across multiple related documents. By identifying when different lexical expressions refer to the same underlying referent, CDCR enables a richer form of content analysis that aggregates information at the level of discourse participants rather than isolated terms. While the ECB+ dataset \cite{cybulska-vossen-2014-using} remains a widely used benchmark for evaluating CDCR models, it primarily emphasizes event resolution and therefore adopts a relatively narrow definition of coreference, i.e., an event is coreferential when it has the same attributes of actor, location, and time \cite{zhukova-etal-2022-towards}. Hence, ECB+ annotates entities as event-dependent and ignores those mentions outside annotated events, thereby preventing the identification of complete actor coverage in the news \cite{zhukova-2022-xcoref}. A narrow definition of coreference relations falls short when a CDCR model trained on such a dataset is applied to polarized news articles, i.e., articles reporting on the same event but presenting different perspectives, emphasizing distinct aspects of the story, and employing markedly distinct vocabulary \cite{zhukova-2022-xcoref}.

Applying CDCR in the news domain for content analysis and to detect linguistic patterns in media coverage requires a reconsideration of what it means for two mentions to refer to the same event or entity. Traditional coreference models are designed to capture strict referential equivalence (e.g., ``Angela Merkel'' - ``the German chancellor''), yet journalistic discourse often employs ``looser'' coreference relations, e.g., near-identity \cite{recasens-etal-2010-typology}, quasi-identity \cite{hovy-etal-2013-events}, euphemisms \cite{gavidia-etal-2022-cats}, metaphors \cite{joseph-etal-2023-newsmet, radford-2020-seeing}, or paraphrases \cite{wahle-etal-2023-paraphrase} that blur referential boundaries \cite{heuss2014comparison}. To address this problem, Hamborg et al. \cite{Hamborg2019a} proposed NewsWCL50, a dataset that annotated concepts prone to bias by word choice and labeled terms such as ``migrants'', ``caravan'', ``threat'', and ``asylum seekers'' as referring to the same social group. Capturing the wide variety of word choices provided a more realistic setting for CDCR models and enabled the analysis of how referents are framed, evaluated, and transformed across contexts \cite{hamborg-donnay-2021-newsmtsc}. 

While NewsWCL50 captures bias through word choice and labeling instances, the annotation scheme only identifies comparably broad concepts more suitable for content analysis than coreference resolution \cite{schreier2012}. This paper proposes a revised annotation scheme for NewsWCL50, which, first, explicitly treats coreference chains as discourse elements (DEs) and conceptual units of analysis, and second, adheres to the definitions and requirements of identity and near-identity relations, providing more fine-grained entities, events, and concepts. We evaluate the annotation scheme by reannotating the entire NewsWCL50 and a subset of ECB+ using the same codebook, and comparing these new datasets using lexical diversity metrics and performance on the same-head-lemma baseline.  Our experiments show that the reannotated datasets NewsWCL50\textsubscript{r} and ECB+\textsubscript{r} exhibit consistently similar metric profiles that lie between those of the original NewsWCL50 and ECB+ datasets. This effectively balances the two codebooks, providing annotations for coreference chains that adhere to conventional definitions of coreference while also addressing the necessity for high lexical diversity in coreferential mentions. The codebook, annotation files in MAXQDA, and the final annotations are available at \url{https://github.com/anastasia-zhukova/NewsWCL50r}.

\section{Related work}

Developing datasets for coreference resolution that capture varying levels of complexity and lexical diversity has been a long-standing research focus in within-document coreference resolution, for example, in \cite{weischedel2011ontonotes, ogorman-etal-2016-richer, mitamura2017events, mitamura2015overview, linguistic2008ace, linguistic2005ace, hovy-etal-2013-events, recasens-etal-2012-annotating}. However, annotation schemes and datasets that systematically explore semantic relations and lexical diversity in CDCR remain comparatively rare.

Ahmed et al. \cite{ahmed-etal-2024-generating} addressed the limited lexical diversity of the ECB+ dataset by generating metaphoric paraphrases for event triggers using GPT-4. Although their approach was applied exclusively to events, it demonstrated that introducing paraphrastic variation can effectively increase lexical diversity and, consequently, the complexity of the CDCR task.

As part of the experiments, we extend this idea by reannotating ECB+ to show that lexical diversity can be enriched not only in events but also in entities. We demonstrate that the diverse lexical choices used to refer to a given entity are often present in existing news texts and tend to depend on how an article seeks to frame or portray that entity. However, capturing those diverse entity mentions requires lowering the degree of coreference identity between mentions.

\section{Lexically-Rich CDCR Annotation}
\label{sec:annotat}

We propose a reannotation scheme that addresses two key limitations of existing datasets: the overly strict identity relations in ECB+ and the overly broad annotation guidelines in NewsWCL50, which result in discourse elements (DEs) that are either too narrowly or too broadly defined. Our approach integrates and refines annotation rules from both schemes, resulting in a framework that is both concept-centric and fine-grained in nature. This section introduces the core terminology and guiding principles of the proposed annotation scheme. The accompanying codebook in the project repository provides detailed instructions and practical annotation examples.

% \subsection{Topology}

\textbf{Mention:} We define \textit{a mention} in a text as a \textit{single word} or \textit{multi-word phrase}, i.e., heads and the head modifiers such as adjectives, compounds, adverbs, direct objects, etc., that trigger a coreference chain. Examples: ``an anxious and uncertain President Trump,'' ``a difficult emotional decision for the president,'' ``less-educated, native-born Americans,'' ``a cruel action.'' We annotate only named and nominal mentions \cite{linguistic2008ace}, and unlike ECB+, we annotate a full noun phrase following the maximum span annotation principle \cite{zhukova-etal-2022-towards}.

\textbf{Coreference chain as a discourse element (DE):} We define \textit{a coreference chain} in the linguistic sense as a single, language-independent, semantic unit to be annotated from a set of related articles, i.e., a discourse element. In a document, a chain is represented by a term or a collection of coreferential terms. For example, ``the current President of the United States,'' which can also be referred to as ``Trump,'' ``President Trump,'' ``the President,'' ``Donald Trump,'' etc. In this study, coreferential chains are usually: (1) \textit{entities}, e.g., actors, organizations, geo-political entities (GPEs), locations, objects \cite{linguistic2008ace, weischedel2011ontonotes}; (2) \textit{events}, such as short-term actions ``deport'' or ``overstep'', or actions or processes with a longer duration, e.g., a meeting, a war, or immigration \cite{linguistic2005ace}; (3)  \textit{concepts}, i.e., frequently covered yet more broadly defined as story elements that aim to aggregate more abstractly related mentions, e.g., a reaction to one or more events, other consequences thereof, or chains of small but interrelated events \cite{schreier2012}.

\begin{table*}
\begin{tabular}{p{0.01\textwidth}|p{0.01\textwidth}|p{0.9\textwidth}}
% \begin{tabular*}{\textwidth}{c|c|l}
% \begin{tabular}{l|l |l }
\hline
   & &\textbf{Annotations}\\
\hline
\makecell[c]{\multirow{5}{*}{\rotatebox[origin=c]{90}{\textbf{COUNTRY: USA}}}} & \makecell[c]{\rotatebox[origin=c]{90}{\textbf{NewsWCL50}}}                  & United States; U.S.; White House; CIA director; Matthew Pottinger; Larry Kudlow; chief White House correspondent; American officials; Washington; personal lawyer; C.I.A.; senior director for Asia at the National Security Council; C.I.A. director; U.S. Senators; CIA director and secretary of State-designate; Senate Foreign Relations Committee; director of the East Asia nonproliferation program at the Middlebury Institute; southern White House
Director of Trump's National Economic Council; U.S. officials; F.B.I. director; 
United Nations Command; His administration; Robert Mueller; Jeffrey Lewis; senators; Kudlow; senior vice president at the Center for Strategic and International Studies; Center for Strategic \& International Studies; Special Counsel; Central Intelligence Agency Director; U.S. diplomats; American troops; U.S. secretary of state; United States government; U.S.-led forces; Mr. Trump’s chief economic adviser; CIA chief; National Security Council; Americans; America; Michael D. Cohen; Mr. Kudlow
 \\ 
                    \cline{2-3} 
                   &  \multirow{4}{*}{\rotatebox[origin=c]{90}{\textbf{NewsWCL50\textsubscript{r}}}}                & 
                        \textit{USA}: United States; U.S.; White House; his administration; Washington; the U.S.-led forces in the conflict; United Nations Command; the White House press office; the Senate; the U.S.; the Senate Foreign Relations Committee; the Washington-based Center for Strategic \& International Studies; the State Department; C.I.A.; signers to the armistice; the government; United States government; Senate Foreign Relations Committee; America; the two countries; the U.N. Command \\
                       \cline{3-3} 
                       &  & \textit{USA\textbackslash Kudlow}: Mr. Kudlow; Larry Kudlow, Mr. Trump's chief economic adviser; Larry Kudlow, director of Trump's National Economic Council\\
                       \cline{3-3} 
                       &  & \textit{USA\textbackslash Pottinger}: Matthew Pottinger, the senior director for Asia at the National Security Council; Matthew Pottinger, senior director for Asian affairs for Trump's National Security Council \\ 
                       \cline{3-3} 
                       &  & \textit{USA\textbackslash Tillerson}: Tillerson; Rex Tillerson  \\ 
                    
\hline

\multirow{6}{*}{\rotatebox[origin=c]{90}{\textbf{GROUP: Suffered people    }}} & \multirow{5}{*}{\rotatebox[origin=c]{90}{\textbf{ECB+}}} & 
                        \textit{t37\_victims\_of\_quake}: 24 people; Five; five people; one person; three people \\ \cline{3-3} 
                       &  & \textit{t37\_people\_ran\_houses}: people  \\ 
                       \cline{3-3} 
                       &  & \textit{t37\_child\_killed}: 1; child; one; one persona   \\ \cline{3-3} 
                       &  & \textit{t37\_50\_people\_injured}: 50; 200; 50 people; dozens; dozens of villagers; five people; hundreds \\ 
                       \cline{3-3} 
                       &  & \textit{t37\_2ppl\_missing}: two other \\ 
                       \cline{2-3} 
                       &  \makecell[c]{\rotatebox[origin=c]{90}{\textbf{ECB+\textsubscript{r}}}}                & 14; 140; 1 dead; 10 people; 14 others; 230,000 people; 24 dead; 24 people dead; 249 people injured; 70 others; a child; a child who died when a wall collapsed; A man; an estimated 14 children still trapped under the rubble; Another four people; around 30 people seriously injured; around 50 people with injuries; around 50 people with injuries sustained when the walls of their houses collapsed; at least 24 people; At least five people; at least one person; death toll; dozens; dozens injured; Dozens of people; dozens of villagers; Five dead; Five people; Four other people; hundreds more injured; Injured people; Many people; more than 1,000 people; more than 200; more than 200 people; one man; One of the fatalities; over 200 injured; seven; six children; some 230,000 people around the Indian Ocean; some people; the children still trapped after the mosque collapse in Blang Mancung village; Twelve people; two others missing; two people  \\ 
\hline

\makecell[c]{\multirow{2}{*}{\rotatebox[origin=c]{90}{\textbf{ACTOR: Warren Jeffs}}}} & \makecell[c]{\rotatebox[origin=c]{90}{\textbf{ECB+}}}                  & \textit{t36\_warren\_jeffs}: attorney; FLDS leader's; he; head; him; his; Jeffs; leader; leader Warren Jeffs; pedophile; Polygamist; polygamist leader Warren Jeffs; Polygamist prophet Warren Jeffs; polygamist sect leader Warren Jeffs; Polygamist Warren Jeffs; Warren Jeffs; Warren Jeffs, Polygamist Leader; who \\ 
                    \cline{2-3} 
                   & \makecell[c]{\rotatebox[origin=c]{90}{\textbf{ECB+r}}}                &  a handful from day one; a problem; a victim of religious persecution; an accomplice for his role; an accomplice to rape by performing a marriage involving an underage girl; an accomplice to sexual conduct with a minor; an accomplice to sexual misconduct with minors; an accomplice to the rape of a 14-year-old girl; FLDS prophet Warren Jeffs; God’s spokesman on earth; her father; his client; Jeffs;  Jeffs, 54; Jeffs, who acted as his own attorney; Jeffs, who was indicted more than two years ago; Mr. Jeffs; one individual, Warren Steed Jeffs; one of the most wicked men on the face of the earth since the days of Father Adam; penitent; Polygamist prophet Warren Jeffs; polygamist sect leader Warren Jeffs; polygamist Warren Jeffs; president; prophet of the Fundamentalist Church of the Jesus Christ of the Latter Day Saints; prophet Warren Jeffs; stone-faced; The 54-year-old Jeffs; the defendant; the ecclesiastical head of the Fundamentalist Church of Jesus Christ of Latter Day Saints; the father of a 15-year-old FLSD member 's child; the highest-profile defendant; the prophet; the self-styled prophet; their client; their spiritual leader; This individual; Warren Jeffs; Warren Jeffs, leader of the Fundamentalist Church of Jesus Christ of Latter Day Saints; Warren Jeffs, polygamist leader \\ 
\hline

\end{tabular}
\caption{The annotation example demonstrating the difference between the original annotation schemes of NewsWCL50 and ECB+ to the proposed lexically diverse yet fine-grained annotation scheme. }
\label{tab:annot_example}
\end{table*}

\textbf{Coreference relations}: To capture mentions referring to concepts that vary in semantic complexity and lexical diversity, we define several types of coreferential relations. Each concept is composed of mentions linked by one or more of these relation types. Importantly, every concept must represent a semantically independent element within the news story, ensuring that it can stand as a distinct unit of meaning in the broader discourse.

\textit{Identity relations}: Identity relations denote a clear equivalence between two mentions that refer to the same concept, such as ``Donald Trump'' and ``the President'' \cite{weischedel2011ontonotes}. These relations may occur between noun phrases (NPs) and verb phrases (VPs), for example, ``met with Donald Trump'' and ``the planned meeting with Donald.'' Synonyms grounded in common knowledge or contextual interpretation also constitute identity relations, such as ``talked about'' – ``discussed,'' ``DACA recipients'' – ``undocumented children,'' or ``Olympics'' – ``sport competition.'' While appositional modifiers like ``John, an artist'' are typically annotated as attributes \cite{weischedel2011ontonotes} or less strict identity relations \cite{ogorman-etal-2016-richer}, in the context of related news articles, such modifiers should be annotated together with their heads, consistent with other modifiers, for instance, the adjectival modifier in ``undocumented children.''

\textit{Near-identity or bridging relations}: To capture relations beyond strict coreference, we adopt the concept of near-identity relations proposed by \cite{recasens-etal-2010-typology}, quasi-identity from \cite{hovy-etal-2013-events}, and bridging relations as defined by \cite{ogorman-etal-2016-richer}. While such relations are typically omitted in annotation schemes like OntoNotes \cite{weischedel2011ontonotes} or ECB+ \cite{cybulska-vossen-2014-using}, we include several types of semantically related but non-identical connections between mentions.
(1) Part–whole (metonymic or meronymic) relations, where a mention represents a part of a larger concept, e.g., ``the Kremlin'' – ``the Russian government'' \cite{recasens-etal-2010-typology, ogorman-etal-2016-richer}.
(2) Context-dependent equivalences between words with distinct literal meanings that refer to the same concept in context, often conveying evaluative judgment, e.g., ``invade'' – ``cross the border'' \cite{Hamborg2019a}. These equivalences also include euphemisms \cite{gavidia-etal-2022-cats} and metaphors \cite{joseph-etal-2023-newsmet}.
(3) Copular constructions formed by verbs such as ``be,'' ``seem,'' or ``feel,'' which associate subjects with attributes, e.g., ``This meeting is a big step forward,'' where ``is'' links ``meeting'' and ``a big step forward'' \cite{wedding2002social, cap2006metaphor}.
(4) Labeling or `calling' relations, introduced by verbs like ``call,'' ``name,'' ``describe,'' or ``denounce,'' which assign evaluative or ideological labels \cite{kosloff2010smearing}. For example, in ``Trump called Kim Jong Un a Rocket Man,'' the pair ``Kim Jong Un'' – ``a Rocket Man'' forms a bridging relation. Similarly, in ``Khan said that Trump behaved like a 12-year-old,'' the pair ``Trump'' – ``a 12-year-old'' should be annotated. 

\textit{Aggregating relations}: A single article or a collection of related articles may include mentions connected through structural relations such as set–subset–element or whole–part relations \cite{clark1975, ogorman-etal-2016-richer}. A typical example is the set–element relation, as in ``three women'' – ``one of these women.'' In some cases, the set itself is abstract, implicit, or missing entirely, leaving only the mentions of the elements. For instance, in the Comey memos, i.e., a series of documents describing Comey’s interactions with Donald Trump, mentions such as ``dinners,'' ``meetings,'' ``encounters,'' and ``conversations with Trump'' represent elements of a broader but unexpressed set of concepts, namely ``Interactions with Trump.'' Concepts composed solely of element mentions (whether noun or verb phrases) are generally more difficult to identify, as their recognition requires a higher level of abstraction, particularly when the overarching set is not explicitly stated in the text.

\textbf{DE types:} The proposed annotation scheme distinguishes ten DE types to capture semantic diversity and discourse structure across documents. ACTION refers to an activity performed by an actor or another DE \cite{linguistic2005ace}, such as ``Negotiate about the peace'' by Trump and Korean officials. At the same time, ACTOR denotes a person proper noun performing an action \cite{linguistic2008ace}, for example, ``Mike Pompeo.'' COUNTRY represents a geopolitical entity (GE) or its institutions \cite{linguistic2008ace}, such as ``the United States of America (USA).'' ACTOR-G is a DE that represents a group of people associated with a country. If DE is labeled as [COUNTRYNAME]-I, it designates individuals who officially represent a country or organization, e.g., ``Korean envoys.'' In contrast, a GE labeled as [COUNTRYNAME]-MISC identifies passive membership or population mentions, such as ``Iranian citizens.'' EVENT captures ongoing or extended activities \cite{linguistic2005ace}, like an ``armed confrontation,'' and GROUP refers to collectives acting together \cite{cybulska-vossen-2013-semantic, cybulska-vossen-2014-using}, such as `demonstrators.'' OBJECT designates non-animated yet salient items \cite{cybulska-vossen-2013-semantic, cybulska-vossen-2014-using}, e.g., ``DNC servers,'' while ORGANIZATION denotes formal non-government entities \cite{linguistic2008ace}, such as ``Wikileaks.'' Finally, MISC encompasses abstract or aggregating concepts \cite{schreier2012}, for example, ``Denuclearization,''ß which unify semantically related mentions that lack an explicit set reference in text.

\begin{table*}[]
\centering
\begin{tabular}{|l|r|r|r|r|r|r|r|r|r|r|r|r|}
\hline
\multirow{2}{*}{\textbf{Dataset}} & \multicolumn{5}{c|}{\textbf{Coreference chains / Discourse elements (DEs)}} & \multicolumn{4}{c|}{\textbf{Mentions}} & \multicolumn{3}{c|}{\textbf{Lexical diversity}} \\
\cline{2-13}
 & all & entity & events & singletons & avg. size & all & entity & event & avg. per doc & UL & PD & MTLD \\
\hline
NewsWCL50 & 134 & 96 & 38 & 4 & 38.2 & 5115 & 3886 & 1229 & 102.3 & 10.71 & 8.99 & 14.54 \\
NewsWCL50\textsubscript{r} & 433 & 374 & 59 & 102 & 15.1 & 6531 & 4758 & 1773 & 130.6 & 6.05 & 9.06 & 15.58 \\
\hdashline
ECB+* & 171 & 112 & 59 & 59 & 2.4 & 407 & 246 & 161 & 16.3 & 2.19 & 1.99 & 4.55 \\
ECB+METAm* & 168 & 109 & 59 & 57 & 2.4 & 399 & 240 & 159 & 16.0 & 2.92 & 3.25 & 6.82 \\
ECB+\textsubscript{r}* & 97 & 84 & 13 & 25 & 14.7 & 1427 & 958 & 469 & 57.1 & 5.88 & 9.04 & 20.65 \\
\hline
\end{tabular}
\caption{General statistics and lexical diversity metrics for the original and reannotated versions of NewsWCL50 and ECB+. An asterisk (*) denotes that a subset of five subtopics was used. }
\label{tab:general}
\end{table*}

\section{Experiments}

\subsection{Dataset reannotation}
We reannotated NewsWCL50 and ECB+ using the proposed annotation scheme to compare the characteristics of the resulting datasets. The objective was to create datasets that, despite their differing topical compositions, produce annotated mentions and DEs with comparable distributions in lexical diversity and dataset complexity for CDCR modeling. For instance, although NewsWCL50 primarily contains political news and ECB+ consists mainly of general or human-interest news, the dataset properties of the reannotated versions should become more balanced compared to the significant discrepancies in the original datasets \cite{zhukova-etal-2022-towards}.

The reannotated version of NewsWCL50, referred to as \textit{NewsWCL50\textsubscript{r}}, aimed to produce more precisely defined coreference chains and minimize annotation ambiguity in the original dataset. First, we divided overly broad concepts into multiple, more specific DEs, ensuring that mentions within a DE share coreferential, meronymic, metonymic, or part–whole relations, and that each entity’s mentions belong exclusively to a single DE. Second, we annotated previously missing mentions and expanded phrases by including non-annotated noun or verb modifiers. Finally, we added small or previously missed concepts, including singleton GEs. We have reannotated the entire corpus of NewsWCL50 \footnote{To ensure that NewsWCL50\textsubscript{r} can serve as a reliable dataset for validating and testing CDCR models (e.g., \cite{bugert-gurevych-2021-event}), we designate topics 0–3 for validation and topics 4–9 for testing.}. 

The reannotation of the ECB+ subset, referred to as \textit{ECB+\textsubscript{r}}, aimed to achieve a complementary objective of producing more loosely defined, lexically diverse DEs. Specifically, the process (1) treated the annotation of events and entities independently, prioritizing frequency of occurrence over adherence to the event–attribute framework, and (2) expanded the annotation scope from minimum span to maximum span to capture the full range of lexical variation present in the text. ECB+\textsubscript{r} consists of five topics from the original test set, which included one political topic and four non-political topics, allowing us to capture diverse lexical choices across both these domains. Specifically, we selected the ``36ecbplus,'' ``37ecbplus,'' ``38ecbplus,'' ``39ecbplus,'' and ``41ecbplus'' topics, and within each topic, we annotated five articles of comparable length to those in NewsWCL50. 

To validate the proposed annotation scheme, we manually curated a randomly selected subset of annotated DEs, i.e., one concept per DE type \cite{steinberger2017large}. The precision of the manual curation for NewsWCL50\textsubscript{r} and ECB+\textsubscript{r} was close to 0.98. Although complete inter-coder agreement would have strengthened the evaluation, the detailed comparison data analysis of both reannotated datasets below provides a robust complementary validation. 

\Cref{tab:annot_example} presents representative examples for both datasets. While NewsWCL50\textsubscript{r} yielded more clearly defined DEs (e.g., ``USA'' transformed from one concept to four entities), ECB+\textsubscript{r} resulted in more broadly defined DEs compared to the original strict coreferential chains (e.g., ``Suffered people'') or annotated with more lexical diversity and free from the actor-attribute role (e.g., ``Warren Jeffs'').

\subsection{Data analysis}
The data analysis comprises three components: (1) a comparison of the general statistics of the datasets, (2) an analysis of lexical diversity, and (3) a performance evaluation using a simple CDCR baseline. The objective is to compare the original and reannotated datasets and, additionally, to evaluate ECB+\textsubscript{r} against the ECB+METAm baseline \cite{ahmed-etal-2024-generating}, which enhances lexical diversity in event mentions through GPT-4–based paraphrasing. The statistics presented in this section are calculated based on the six topics/30 documents of the NewsWCL50 test set, as well as the five subtopics/25 documents of ECB+. Consistent with \cite{zhukova-2022-xcoref}, we report the statistics for the original NewsWCL50 excluding the ambiguous DE type ACTOR-I. The following data analysis uses the terms ``coreference chain'' and DEs interchangeably. 

\textbf{General statistics} were assessed based on the number of identified DEs and the number of mentions comprising these DEs. As presented in \Cref{tab:general}, NewsWCL50\textsubscript{r} exhibits a threefold increase in the total number of DEs, accompanied by a 2.5-fold reduction in average chain size, indicating finer-grained and more precise annotation. The total number of mentions increased by 27.7\%, as did the average number of annotated mentions per document, reflecting improved annotation coverage. In contrast, ECB+\textsubscript{r} demonstrates an opposite trend, with the number of DEs decreasing by 1.8 times but the average chain size expanding by 6.1 times. Notably, the mean DE size remains comparable between the two datasets, ranging from 14.7 to 15.1 mentions per chain. Although the total number of annotated mentions in ECB+\textsubscript{r} increased by 3.5 times, the per-document average remains lower than in NewsWCL50\textsubscript{r}, mainly due to the shorter length of ECB+ articles.

\begin{figure}
    \centering
    \includegraphics[width=0.87\linewidth]{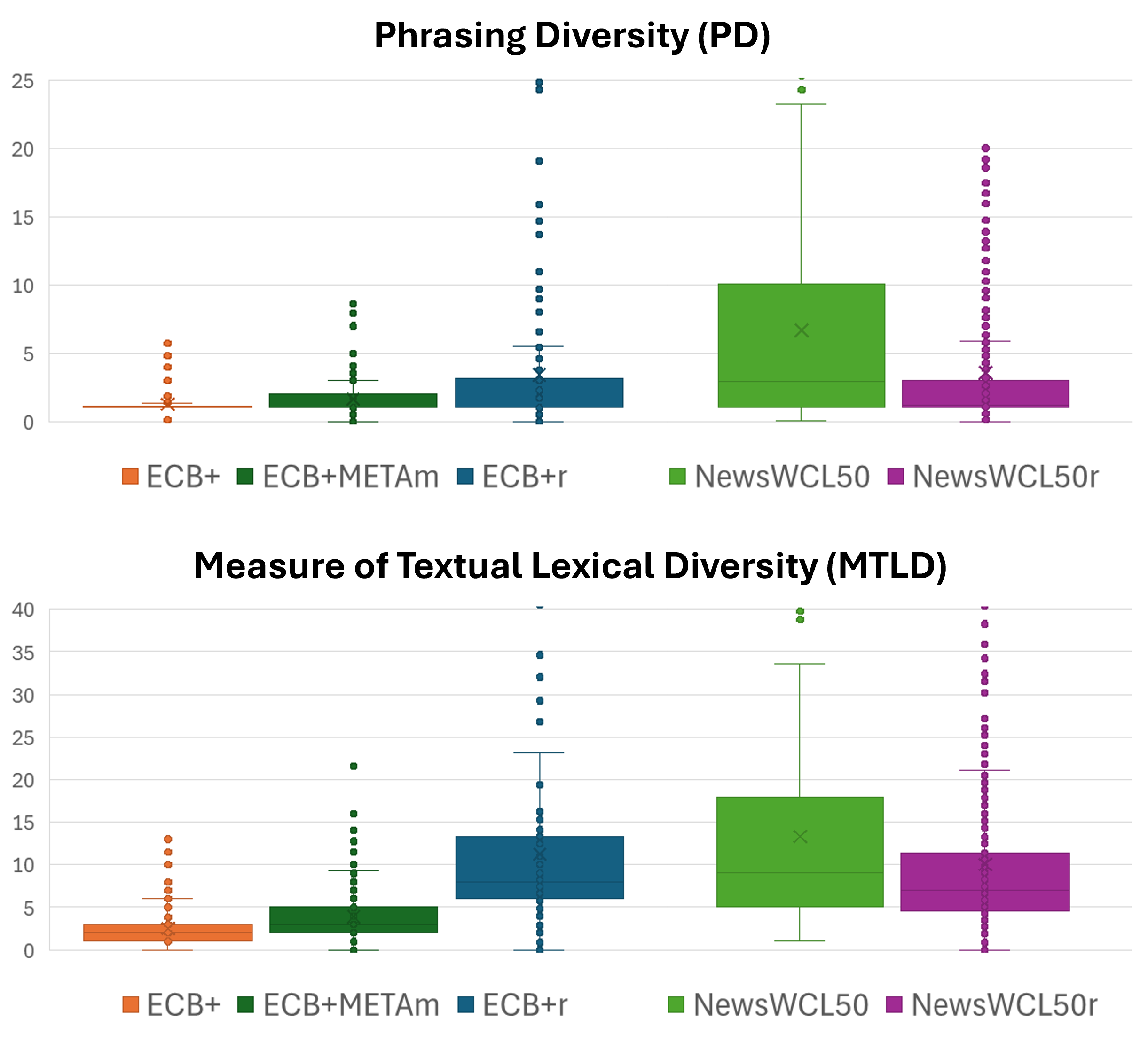}
    \caption{Distribution of lexical diversity measured by PD and MTLD. Both NewsWCL50\textsubscript{r} and ECB+\textsubscript{r} exhibit comparable distributions, demonstrating that the proposed annotation scheme achieves a balanced level of lexical diversity in the coreference chains or DEs across news articles, regardless of domain differences. }
    \label{fig:divesity}
\end{figure}

\textbf{Lexical diversity} is a defining characteristic of CDCR datasets, indicating the extent to which coreferential mentions display paraphrastic variation and semantic flexibility. It is measured using three metrics: (1) the average number of unique head lemmas per coreference chain (UL) \cite{eirew-etal-2021-wec}, (2) the phrasing diversity metric (PD) \cite{zhukova-etal-2022-towards}, and (3) the measure of textual lexical diversity (MTLD) \cite{McCarthy2010, ahmed-etal-2024-generating}. As shown in \Cref{tab:general}, NewsWCL50\textsubscript{r} and ECB+\textsubscript{r} exhibit comparable values for UL and PD. The UL value decreases for NewsWCL50\textsubscript{r}, reflecting finer-grained DE annotation and reduced ambiguity. At the same time, both UL and PD increase for ECB+\textsubscript{r} due to the inclusion of more loosely coreferent mentions. Moreover, ECB+\textsubscript{r} demonstrates a substantial increase in MTLD compared to the ECB+METAm baseline, suggesting that higher lexical diversity among annotated mentions can be achieved from the same non-generated text just by using a different annotation scheme. \Cref{fig:divesity} presents the distributions of PD and MTLD, showing that both reannotated datasets follow similar patterns. This consistency suggests that the proposed annotation scheme yields more balanced coreference chains, thereby avoiding the overly broad annotations of the original NewsWCL50 and the excessively narrow ones of ECB+.

\textbf{Performance analysis} was conducted using a simple same-head-lemma baseline \cite{cybulska-vossen-2014-using}. Model performance was evaluated with the CoNLL F1 score, i.e., a standard metric for coreference resolution that averages the B3, MUC, and CEAF\textsubscript{e} measures (e.g., as applied in \cite{cybulska-vossen-2014-using, bugert-gurevych-2021-event, eirew-etal-2021-wec}). We evaluated non-singleton chains only. As shown in \Cref{tab:performance}, NewsWCL50\textsubscript{r} and ECB+\textsubscript{r} achieve comparable results (54.08 vs. 52.92), representing a more balanced performance relative to the larger discrepancies observed in the original datasets. This outcome indicates that the reannotated datasets provide a moderate level of difficulty for CDCR models, i.e., neither as easily resolvable as the original ECB+ chains nor as challenging due to excessive semantic breadth as in the original NewsWCL50.

The general statistics (\Cref{tab:general}) demonstrate consistent annotation patterns across datasets, while lexical diversity metrics confirm that both corpora achieve balanced variation in wording and semantic richness. Furthermore, comparable model performance scores in the baseline (\Cref{tab:performance}) indicate that the reannotated datasets yield a moderate and comparable level of difficulty for CDCR models. Together, these findings validate the proposed scheme as both methodologically sound and effective in producing lexically diverse, semantically coherent annotations that are suitable for both NLP and social science research.

\begin{table}[]
\centering
\begin{tabular}{|l|c|c|c|c|}
\hline
\textbf{Dataset} & \textbf{MUC} & \textbf{B3} & \textbf{CEAF\textsubscript{e}} & \textbf{CoNLL} \\
 \hline
NewsWCL50  & 82.49  & 49.70 & 11.96 & 48.05 \\
NewsWCL50\textsubscript{r}  & 79.59  & 51.82  & 30.82 & 54.08 \\
\hdashline
ECB+*  & 68.35  & 73.27  & 68.10 & 69.91 \\
ECB+METAm*  & 49.06  & 67.22 & 57.85 & 58.04 \\
ECB+\textsubscript{r}*  & 82.32  & 47.98  & 28.46 & 52.92 \\
\hline
\end{tabular}
\caption{The performance of the same-head-lemma baseline. }
\label{tab:performance}
\end{table}

\subsection{Discussion}
The proposed annotation scheme focuses on a lexical diversity challenge for CDCR models, requiring them to learn and recognize looser coreference relations that link phrases with diverse word choices \cite{zhukova-etal-2022-towards}. By incorporating paraphrases, metonymic relations, euphemisms, metaphors, and evaluative wording, the dataset compels models to move towards capturing deeper semantic and contextual equivalences. This not only tests model robustness but also encourages the development of CDCR systems capable of handling the linguistic variability typical of real-world news discourse, particularly in the study of media bias, framing, and discourse \cite{zhukova-2022-xcoref}. This capability bridges computational methods with critical approaches in media and communication studies, enabling large-scale content analysis that captures not only what is being discussed, but also how and why it is framed in particular ways \cite{hamborg-donnay-2021-newsmtsc}.

\section{Conclusion}
The proposed annotation scheme advances CDCR by integrating lexical diversity and looser identity relations into a consistent and balanced annotation framework. By refining event and entity boundaries while preserving semantic variability, it produces datasets that better reflect the linguistic complexity of real-world news discourse. This approach not only presents a meaningful challenge for CDCR models, requiring them to recognize paraphrastic and context-dependent relations, but also enables large-scale, data-driven analyses of media bias, framing, and discourse.

\bibliographystyle{IEEEtran}  
\bibliography{bibliography}

\end{document}